\documentclass[letterpaper, 10 pt, conference]{ieeeconf}  % Comment this line out if you need a4paper

\IEEEoverridecommandlockouts                              % This command is only needed if 
\usepackage[utf8]{inputenc}
\usepackage{multicol}
\usepackage{amssymb,graphicx,epstopdf,amsmath,color,tabularx,float,url,bm}
\usepackage[noadjust]{cite} % groups citations
\usepackage{subcaption,overpic}
\usepackage{placeins}
\usepackage{hyperref,dblfloatfix,verbatim}
\usepackage{cleveref}
\usepackage{multirow}
\usepackage{hhline}
\usepackage{tikz}
\usepackage{pict2e}
\usepackage[T1]{fontenc} % Fixes the issue of £ showing up as $
\usepackage{gensymb}
\usepackage[ruled,vlined]{algorithm2e}

\usepackage{array}

\usepackage{tikz}
\usetikzlibrary{shapes,arrows}
\tikzstyle{decision} = [diamond, draw, fill=blue!20, 
    text width=4.5em, text badly centered, node distance=3cm, inner sep=0pt]
\tikzstyle{block} = [rectangle, draw, fill=blue!20, 
    text width=5em, text centered, rounded corners, minimum height=4em]
\tikzstyle{line} = [draw, -latex']
\tikzstyle{cloud} = [draw, ellipse,fill=red!20, node distance=3cm,
    minimum height=2em]

\makeatletter
\newcommand{\thickhline}{%
    \noalign {\ifnum 0=`}\fi \hrule height 1pt
    \futurelet \reserved@a \@xhline
}
\newcolumntype{"}{@{\hskip\tabcolsep\vrule width 1pt\hskip\tabcolsep}}
\makeatother

\title{\LARGE \bf
A Biomimetic Tactile Fingerprint Induces Incipient Slip
}

\author{Jasper W. James$^{1,2}$, Stephen J. Redmond$^{3}$ and Nathan F. Lepora$^{1,2}$% <-this % stops a space
\thanks{JJ was
supported by the EPSRC CDT in Future Autonomous and Robotic Systems (FARSCOPE). SJR was supported by the Science Foundation Ireland (grant 17/FRL/4832). NL was supported by a Leadership Award from the Leverhulme Trust on `A biomimetic forebrain for robot touch' (RL-2016-39).}% <-this % stops a space
\thanks{$^{1,2}$JJ and NL are with the Dept. of Engineering Mathematics,
        University of Bristol, Bristol, UK and The Bristol Robotics Laboratory, Bristol, UK.
        {\tt\small \{jj16883, n.lepora\}@bristol.ac.uk}}%
\thanks{$^{3}$SJR is with the School of Electrical and Electronic Engineering at University College Dublin, Dublin, Ireland. 
        {\tt\small stephen.redmond@ucd.ie}}%
\thanks{Accepted into IROS 2020.}
}

\begin{document}
\maketitle
\thispagestyle{empty}
\pagestyle{empty}

%%%%%%%%%%%%%%%%%%%%%%%%%%%%%%%%%%%%%%%%%%%%%%%%%%%%%%%%%%%%%%%%%%%%%%%%%%%%%%%%
\begin{abstract}
We present a modified TacTip biomimetic optical tactile sensor design which demonstrates the ability to induce and detect incipient slip, as confirmed by recording the movement of markers on the sensor's external surface. Incipient slip is defined as slippage of part, but not all, of the contact surface between the sensor and object. The addition of ridges - which mimic the friction ridges in the human fingertip - in a concentric ring pattern allowed for localised shear deformation to occur on the sensor surface for a significant duration prior to the onset of gross slip. By detecting incipient slip we were able to predict when several differently shaped objects were at risk of falling and prevent them from doing so. Detecting incipient slip is useful because a corrective action can be taken before slippage occurs across the entire contact area thus minimising the risk of objects been dropped.
\end{abstract}
%%%%%%%%%%%%%%%%%%%%%%%%%%%%%%%%%%%%%%%%%%%%%%%%%%%%%%%%%%%%%%%%%%%%%%%%%%%%%%%%
\section{Introduction}
For autonomous robots to operate effectively in the world they must be able to handle objects safely. When grasping an object, the robot must find the \textit{Goldilocks} zone between gripping too tightly and damaging the object, and too loosely and dropping the object; both of which could result catastrophic damage to the object. The ability of a tactile sensor to detect object slippage can set a lower bound on the safe gripping force, as once slip is detected the grasp can be tightened and the object prevented from falling. One key focus of slip-detection research in robotics is to find when parts of the sensor begin to slip whilst other parts remain in static contact. This phenomenon of slippage of part, but  not all, of the contact surface(s) between the sensor and object is termed incipient slip \cite{Chen2018}.

Incipient slip is also observed in the skin of the human finger pad. Due to the non-uniform distribution pressure (and hence local normal forces) across the finger pad, some regions may slip before others, where the traction (proportional to normal force) is smaller \cite{Francomano2013ArtificialReview, Tada2004AnFingertip}. On the mechanoreceptor level, this provides significant stimulation of the FA-1 and SA-1 afferents, and has been observed to trigger an increase in grip force \cite{Johansson1987SignalsGrip}, although the exact biomechanical and neural mechanisms underpinning this response have yet to be definitively identified. Detecting incipient slip is useful because corrective action can be taken before slippage occurs across the entire contact area, hence significantly reducing the likelihood of the object being dropped.

\begin{figure}[t]
    \centering
    \begin{subfigure}[b]{0.23\textwidth}
        \centering
        \begin{overpic}[height=4.3cm]{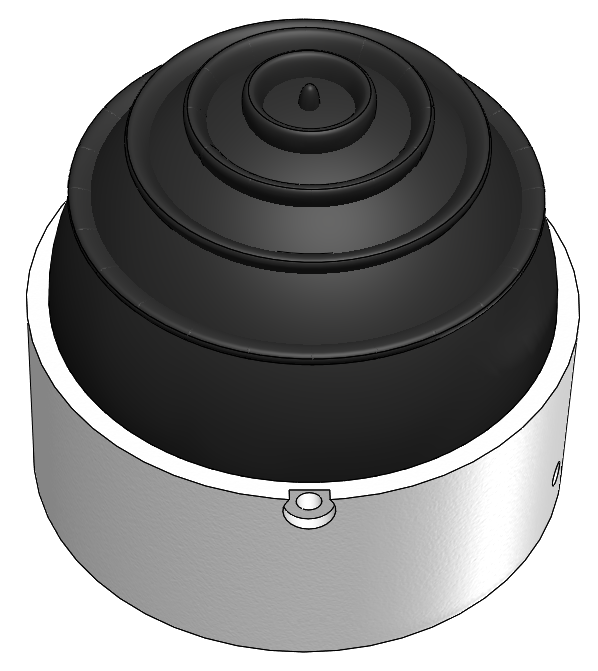}
        \put(0,90){(a)}
        \end{overpic}
    \end{subfigure}
    \begin{subfigure}[b]{0.23\textwidth}
        \centering
        \begin{overpic}[height=4.3cm]{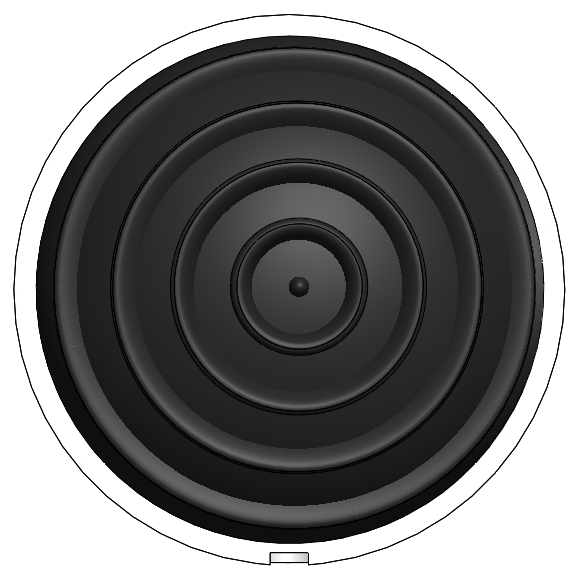}
        \put(0,90){(b)}
        \end{overpic}
    \end{subfigure}
    \begin{subfigure}[b]{0.47\textwidth}
        \centering
        \begin{overpic}[trim={0 1cm 0 0}, clip=true, width=\textwidth]{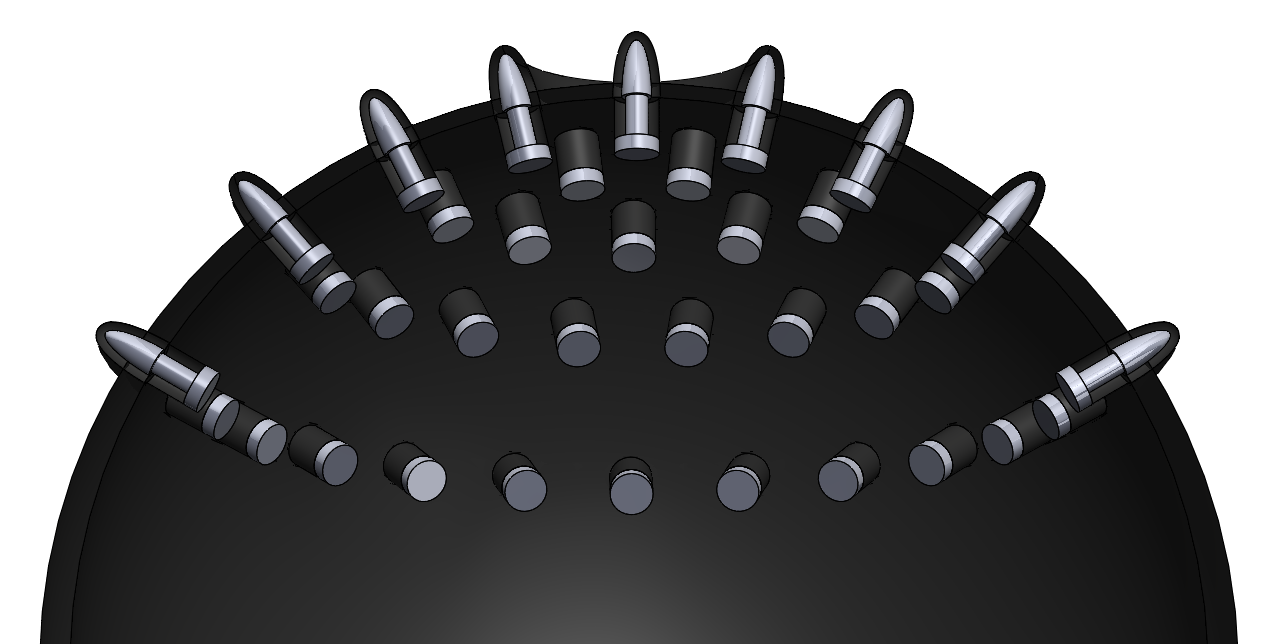}
        \put(0,40){(c)}
        \end{overpic}
    \end{subfigure}
    \caption{(a) \& (b) CAD renderings of new sensor showing the concentric ring pattern of the fingerprint-inspired ridges to facilitate the shear deformation of incipient slip. The ridges extend 2 mm above the surface of the sensor. (c) Cutaway showing how the pins are embedded in the skin.}\label{fig:cads}
\end{figure}

The aim of this study is to develop a sensor which displays incipient slip by mimicking the biomechanical behaviours of the human finger pad. We modify an existing biomimetic optical tactile sensor - the TacTip \cite{ward2018tactip} - which has previously been effective at slip detection, yet does not exhibit incipient slip behavior \cite{James2018SlipSensor}. The TacTip has an internal pins structure which is inspired by the protrusions of the dermal papillae into the epidermis. Here, we extend this mimicry to include ridges on the external surface to facilitate the shear deformation required for incipient slip to occur (Fig. \ref{fig:cads}), as occurs on the human finger pad, facilitated by the fingerprint ridges. \textcolor{black}{This builds upon prior work applying fingerprint mimicry to the TacTip \cite{Cramphorn2017AdditionAcuity} by targeting a specific phenomenon - incipient slip - and designing the sensor such that it provides the necessary physical conditions to induce incipient slip.}

There are significant barriers that make the elicitation and detection of incipient slip challenging. Primarily, when a sensor is deformed by a shear force - as is experienced when holding an object - the surface must have sufficient elasticity to allow parts of the sensor to slip while other parts remain stuck, which requires shear deformation between the slipping and stuck parts of the external sensor surface. Additionally, this must be detectable from the sensor output, in this case the camera inside the sensor. We demonstrate that the modified TacTip presented here displays incipient slip for a substantial duration preceding gross slip, and that the sensor can detect this signal with high accuracy when using a convolutional neural network to interpret the movement of internal pins as measured by the internal sensor camera. 

\section{Background}
Slip detection research can be split into three distinct categories: \textbf{gross} slip, where the entire sensor surface is slipping; \textbf{incipient} slip (investigated here), where parts of the sensor contact with the object is slipping, but not all of it; and slip \textbf{prediction}, where other features of the sensor data can be used to predict when slip is about to occur \cite{Chen2018}. There are many excellent studies involving slip detection; for an in-depth presentation of these methods, we direct the interested reader to reviews by Chen et al. (2018) \cite{Chen2018} and Kappasov et al. (2015) \cite{Kappassov2015TactileReview}. In the following, our focus is limited to prior art relating to incipient slip.
%%%%%%%%%%%%%%%%%%%%%%%%%%%%%%%%%%%%%%%%%%%%%%%%%%%%%
\begin{figure}[t]
    \centering
    \begin{subfigure}[b]{0.23\textwidth}
        \begin{overpic}[width=\textwidth, trim={0 0 0 -1cm}, clip=true]{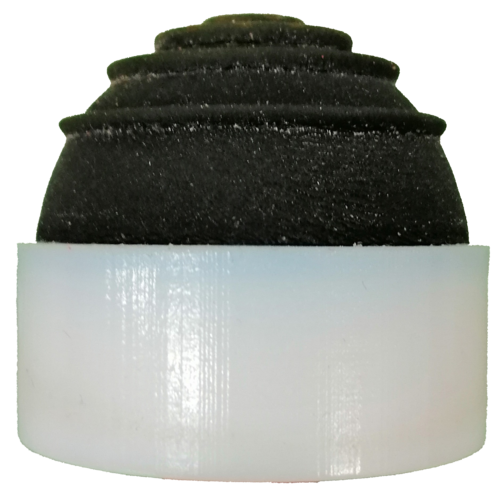}
        \put(50,97){\makebox(0,0){New Sensor (fingerprint)}}
        \put(96,-100){\color{black}\rule{1pt}{235pt}}
        \end{overpic}
    \end{subfigure}
    \begin{subfigure}[b]{0.23\textwidth}
        \begin{overpic}[width=\textwidth, trim={0 0 0 -1.2cm}, clip=true]{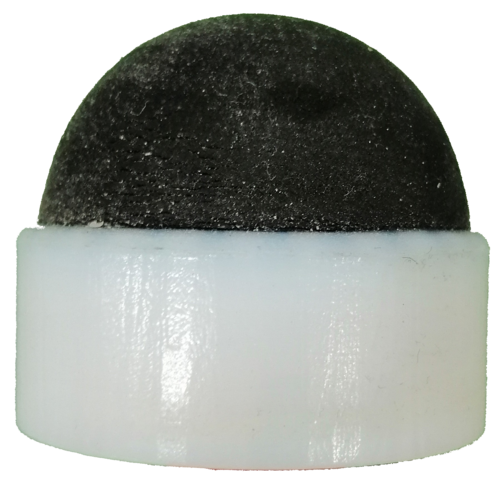}
        \put(50,98){\makebox(0,0){Previous Sensor (smooth)}}
        \end{overpic}
    \end{subfigure}
    \begin{subfigure}[b]{0.23\textwidth}
        \includegraphics[width=\textwidth]{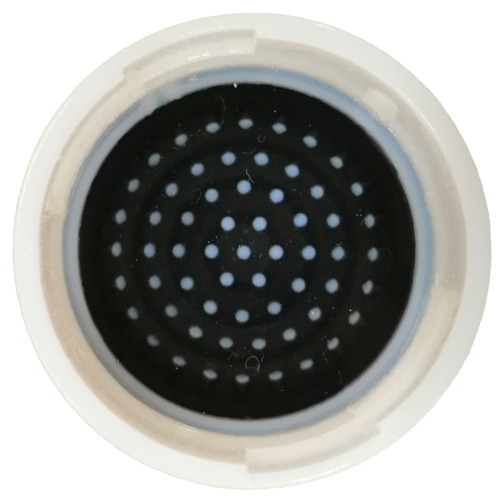}
    \end{subfigure}
    \begin{subfigure}[b]{0.23\textwidth}
        \includegraphics[width=\textwidth]{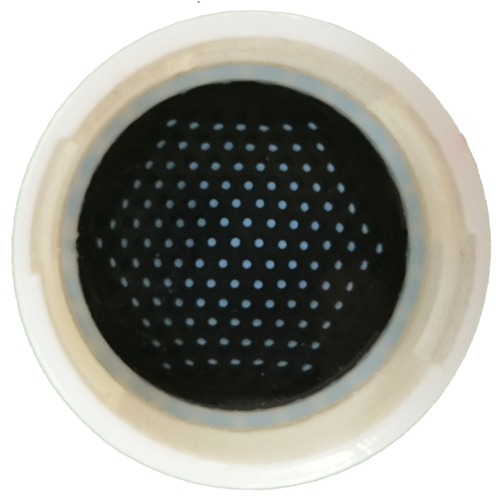}
    \end{subfigure}
    \caption{Comparison of new sensor (left column) and previous TacTip design (right column). The added ridges distort the otherwise uniform surface to facilitate incipient slip. The new sensor also has a smaller pin density.}\label{fig:tips_compare}
\end{figure}
%%%%%%%%%%%%%%%%%%%%%%%%%%%%%%%%%%%%%%%%%%%%%%%%%%%%%%%%%%%%%%%%%%%%%%%%%%%%%%

Dong et al. (2019) were able to detect incipient slip using the GelSlim sensor by observing the movement of markers on the periphery of the contact surface when grasping various objects in a pinch grasp \cite{Dong2019MaintainingSlip}. Su et al. (2015) observed vibration in BioTac sensors to predict slip was going to occur 30 ms before it was detected by an inertial measurement unit atop an object \cite{su2015force}; however, the authors show no evidence that the mechanism allowing them to achieve this detection was indeed incipient slip. Fujimoto et al. (2003) use a PVDF sensor in a ridged skin to detect incipient slip but, again, do not definitively prove that the sensor signal is incipient slip \cite{Fujimoto}. Rigi et al. (2018) used a DAVIS event-based neuromorphic camera to detect incipient slip by reconstructing the surface contact area and detecting when parts of it moved \cite{Rigi2018ADAVIS}. 

A novel sensor for detecting incipient slip was presented by Khamis et al. (2018) \cite{Khamis2018PapillArray:Validation}. The PapillArray consists of a 3x3 array of silicone pillars of differing heights. When compressed and a shear force applied, the pillars further from the centre - which are shorter and therefore under a smaller normal force loading are encouraged to slip before the central pillars. Measuring differing pillar deflection is sufficient to detect incipient slip. The PapillArray pillars can also be used to estimate the coefficient of friction and force, and are highly sensitive to vibration \cite{Khamis2019ASensor}.
%%%%%%%%%%%%%%%%%%%%%%%%%%
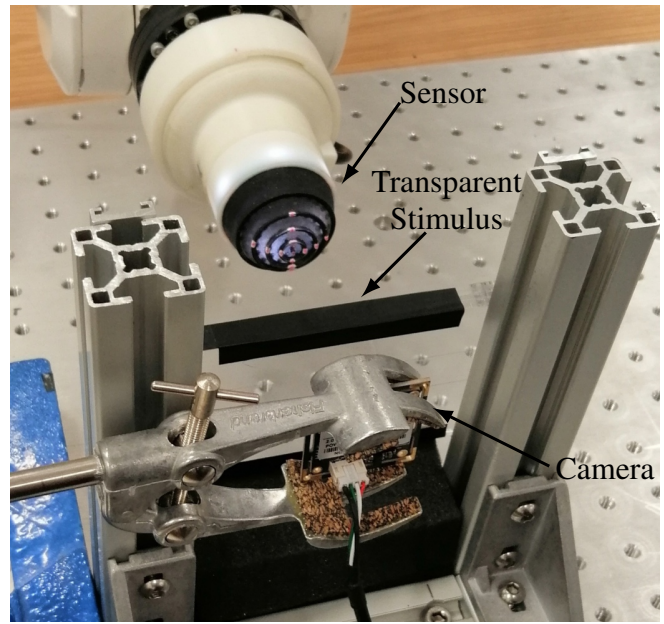
\begin{figure}[t]
    \centering
    \begin{overpic}[trim={0cm 3.5cm 0cm 3cm},clip=true, width=\linewidth]{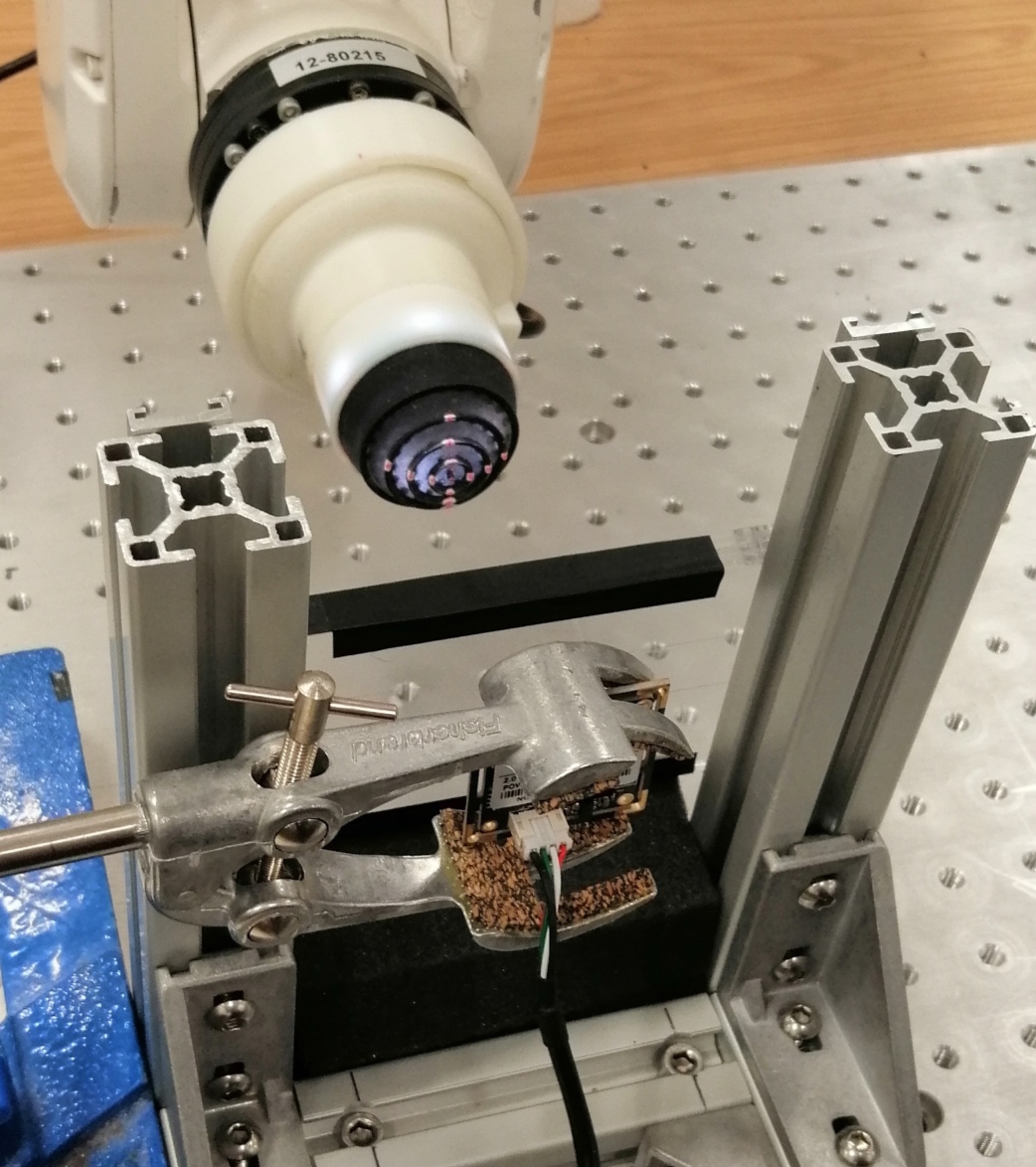}
    \put(60,80){\large Sensor}
    \put(60,80){\linethickness{0.4mm}\vector(-5,-7){9}}
    \put(55,64){\parbox{6em}{\centering\large Transparent Stimulus}}
    % \put(65,60){\linethickness{0.4mm}\vector(0,-1){20}}
    \put(64,60){\linethickness{0.4mm}\vector(-5,-5){10}}
    \put(84,22){\large Camera}
    \put(83,24){\linethickness{0.4mm}\vector(-2,1){18}}
    \end{overpic}
    \caption{Rig for collecting incipient slip data. Modified TacTip attached to an ABB IRB120 six-axis robotic arm; external camera held in a clamp and the transparent acrylic sheet - black tape added for clarity - moving in a low friction rail.}
    \label{fig:rig}
\end{figure}
%%%%%%%%%%%%%%%%%%%%%%%%%%%

This work introduces a modified version of an existing sensor, the TacTip biomimetic optical tactile sensor~\cite{ward2018tactip}. Prior work performed with the TacTip includes gross slip detection using a support vector machine (SVM) \cite{James2018SlipSensor}: objects were allowed to undergo gross slip which had to be detected and the sensor moved forward (increasing the normal force applied) to prevent the object being dropped, catching the objects after slipping by as little as 15 mm. \textcolor{black}{In other work, Cramphorn et al. (2017) explored the effect of a adding protrusions to the outer TacTip surface; however, the aim of that work was to enhance spatial acuity~\cite{Cramphorn2017AdditionAcuity}} not enhance incipient slip. The protrusions also do not resemble a fingerprint so the focus greatly differs from that of this work.

%%%%%%%%%%%%%%%%%%%%%%%%%%%%%%%%%%%%%%%%%%%%%%%%%%%%%
\begin{figure*}[t]
    \centering
    \begin{subfigure}[b]{0.49\textwidth}
        \begin{overpic}[trim={4cm 0cm 4cm 0cm}, clip=true, width=\textwidth]{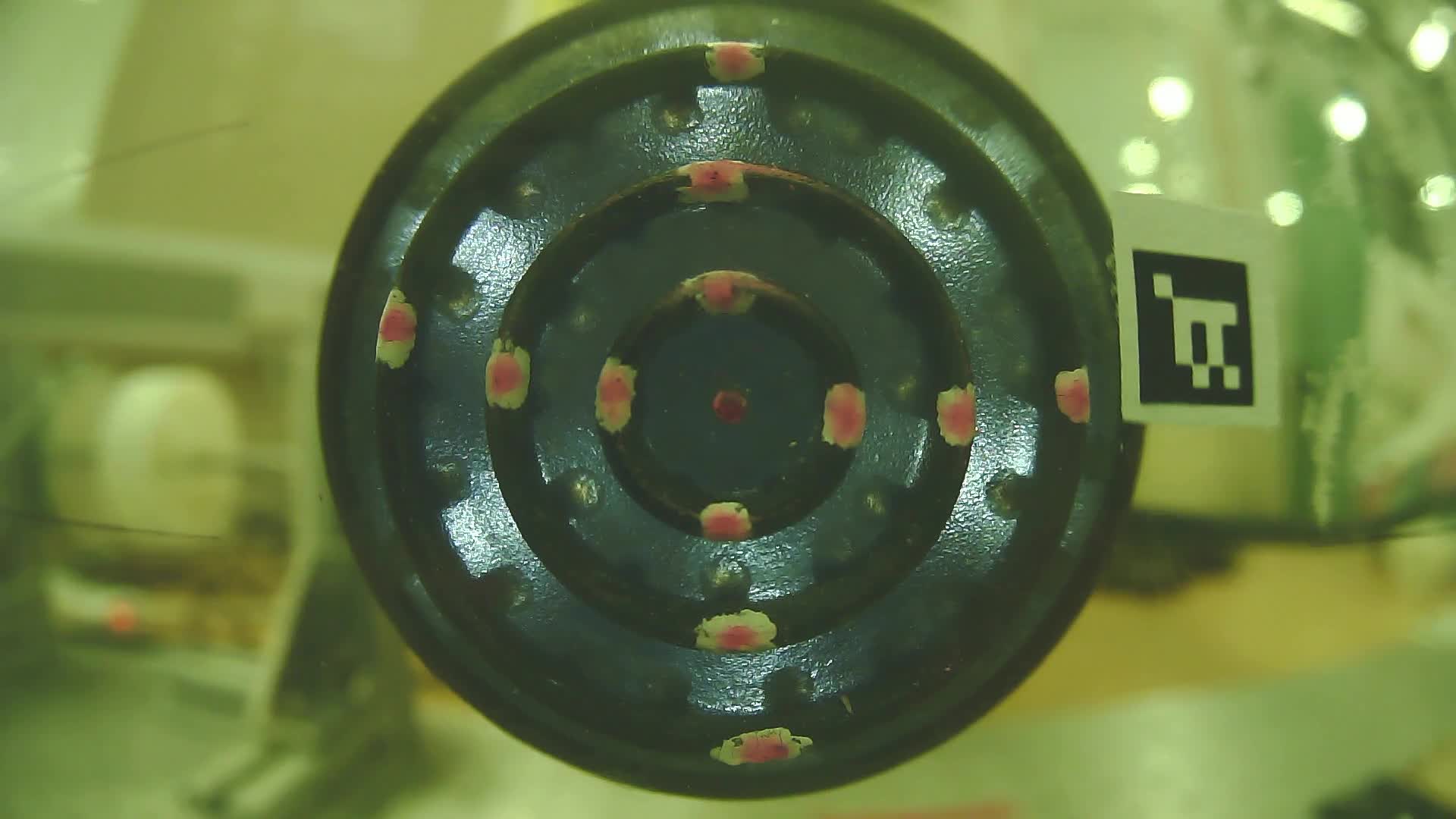}
        \put(0,31.5){\color{white}\rule{84pt}{1pt}}
        \put(0,47.5){\color{orange}\rule{125pt}{1pt}}
        \put(90,43){\color{red}\rule{24pt}{1pt}}
        \end{overpic}
        \caption{Prior to incipient slip}
        \label{fig:run_start}
    \end{subfigure}
    \centering
    \begin{subfigure}[b]{0.49\textwidth}
        \begin{overpic}[trim={4cm 0cm 4cm 0cm},clip=true, width=\textwidth]{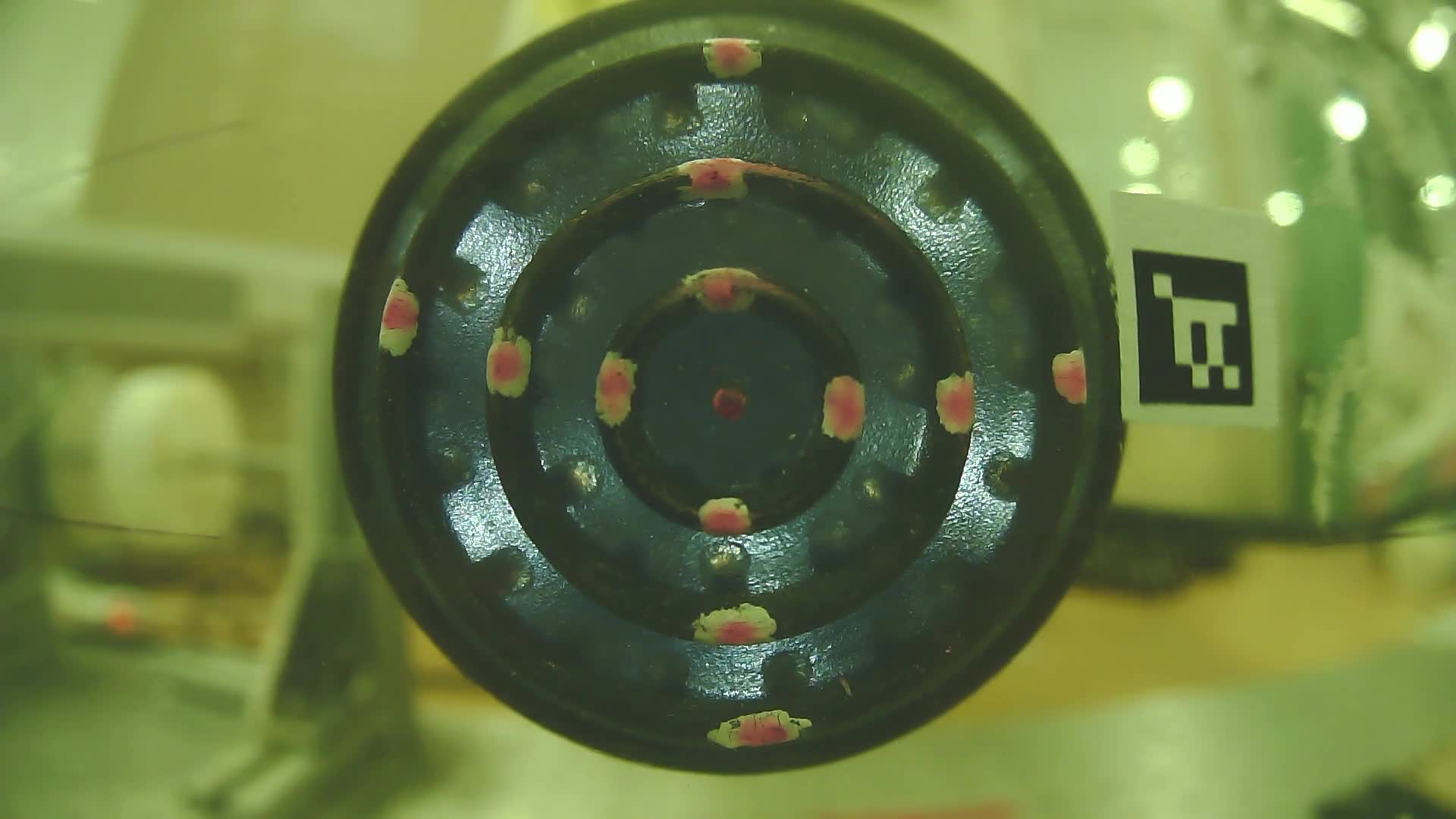}
        \put(0,31.5){\color{white}\rule{84pt}{1pt}}
        \put(0,47.5){\color{orange}\rule{125pt}{1pt}}
        \put(90,43){\color{red}\rule{24pt}{1pt}}
        \put(51,45){\color{orange} Static}
        \put(30,28){\color{white}Slip}
        \end{overpic}
        \caption{Prior to gross slip}
        \label{fig:run_incipient}
    \end{subfigure}
    \caption{External camera view of sensor with markers attached to track fixed points on the surface. (a) Frame from the start of the data collection. (b) Frame taken close to gross slip occurring. For reference, the coloured lines are in the same locations in both images. The marker closest to the white line shows clear movement, however, the marker near the orange line and the ArUco optical marker do not move. Therefore, we see incipient slip: slip of part but not all of the contact surface.}\label{fig:tips_side}
\end{figure*}
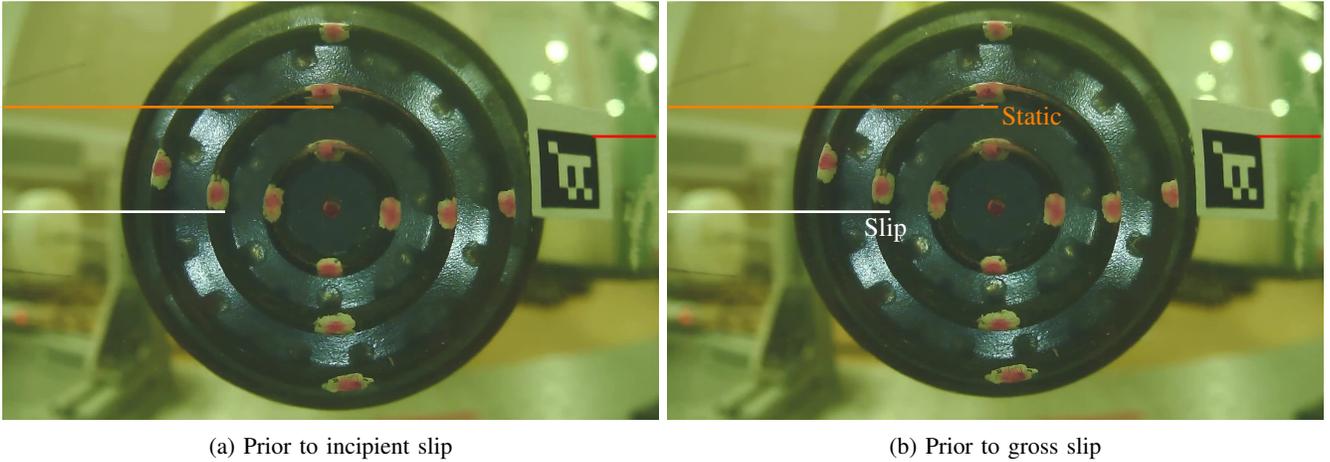
%%%%%%%%%%%%%%%%%%%%%%%%%%%%%%%%%%%%%%%%%%%%%%%%%%%%
\section{Materials and Methods}
\subsection{Modified Sensor Design}
The sensor presented in this work is a modification of the TacTip, a biomimetic optical tactile sensor \cite{ward2018tactip,chorley2009development}. The TacTip contains  127  pins  arranged  in  a  hexagonal pattern on the inside of a hemisphere constructed from TangoBlack+, a rubber-like material. The tip is filled with silicone gel (RTV27905) and sealed with a clear acrylic lens. This makes it compliant but allows it to quickly
reform its shape and reduces hysteresis. A camera (model  ELP-USBFHD01M-L21) is focused on the inside of the sensor and can be used to record the positions of the pins or capture the raw images \cite{Lepora2019FromSensor} as a method of transducing deformation of the external skin surface.

The key functional requirements for the modification are to introduce a traction differential (i.e. different normal force in different places) and an ability for shear displacement to apply to the parts of the sensor (the ridges) that make contact with the object surface, as occurs in the human finger pad \cite{Tada2004AnFingertip} such that some of the contact area slips and some does not. Previously, when an object slipped across the TacTip, there was no discernible incipient slip \cite{James2018SlipSensor}; the pins all moved together in a linear fashion when gross slip occurred. To solve this, we have added five rings of increasing radius, from a central ring of zero radius (in effect an external pin). 

In the TacTip, the pins consist of heads of VeroWhite material atop small pillars of TangoBlack+. Here, we extend the pin-heads such that they protrude through the pillars and into the ridges (Fig \ref{fig:cads}(c)). The pins act as mechanical transducers so any motion of the ridges results in visible motion of the pins from inside the sensor. \textcolor{black}{Cramphorn et al. (2017) utilised this transduction principle to improve the spatial acuity of the sensor by adding small (0.5mm) protrusions to the surface of the TacTip, however their focus was purely on perception not slip \cite{Cramphorn2017AdditionAcuity}}. Here we modify the TacTip with the express aim of allowing it to experience and subsequently transduce incipient slip (Table \ref{tab:problems}).

Firstly, we have added raised ridges which mimic the friction ridges in the dermis of the human fingertip. Under the Coulomb model of friction, the maximum frictional force is proportional to the normal force. Therefore, the addition of raised ridges results in the neighbouring contact areas being spaced apart and subject to notably different normal forces, aided by the curvature of the TacTip exterior. Previously, the smooth hemispherical surface meant that neighbouring regions of contact were subject to very similar normal forces.

Secondly, we have reduced the thickness of the skin between rings from 1 mm to 0.5 mm, to facilitate shear deformation of the sensor contact surface. Even though the ridges will lead to differing normal forces, the rings are still linked by the interstitial skin. The thicker the skin  the higher the coefficient of elasticity, which hinders shear deformation (skin stretch) and hence diminishes the associated incipient slip phenomenon. It should be noted that the added ridges will also bend, which further allows parts of the skin to move independently, and therefore reduces the dependency on the skin stretch for incipient slip to occur.

Finally, as the pins are set in gel, when one pin moves in the TacTip it causes neighbouring pins to move. One consequence of this is that if a single region of the sensor surface was to move it would cause many adjacent pins to move due to this mechanical coupling. To minimise this effect we increase the spacing of the pins such that the projected radius of each ring of pins is 4 mm greater than the ring inside (Fig. \ref{fig:tips_compare} compares the two designs). Previously the spacing was 3 mm between all adjacent pins laid out in a hexagonal pattern. We also increased the thickness of the pins to reduce the potential for damage when removing support material after 3D printing.

\begin{table}[b]
\centering
\begin{tabular}{l l p{4.5cm}}
\hline \hline
 & Modification                      & Benefit             \\ \hline \hline
1 & Add raised ridges   &   Adjacent contact regions have differing normal forces to create traction differential   \\ \hline
2 & Reduce skin thickness  &  Skin stretchier to facilitate incipient slip\\ \hline
3 & Reduce pin density & Decrease coupling of internal pins to facilitate detection of incipient slip\\
\hline \hline
\end{tabular}
\caption{The key modifications to TacTip design to facilitated occurrence and detection of incipient slip.}
\label{tab:problems}
\end{table}
%%%%%%%%%%%%%%%%%%%%%%%%%%%%%%%%%%%%%%%%%%%%%%%%
\begin{figure}[t]
    \centering
    \begin{subfigure}[b]{0.49\textwidth}
        \centering
        \includegraphics[width=\textwidth]{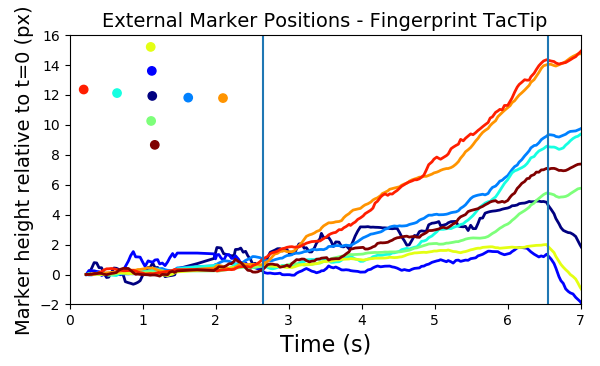}
    \end{subfigure}
    \begin{subfigure}[b]{0.49\textwidth}
        \centering
        \includegraphics[width=\textwidth]{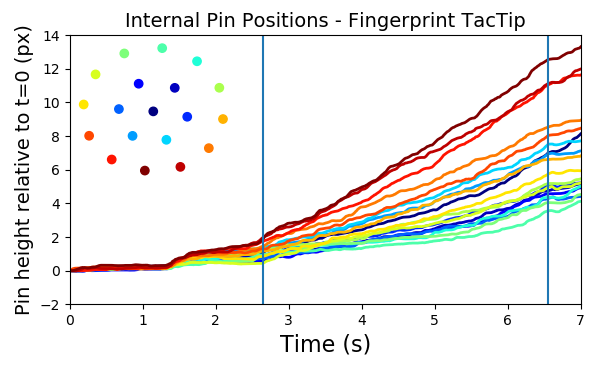}
    \end{subfigure}
    \caption{Top - External markers from newly designed sensor. Blue lines indicate where incipient slip begins (left vertical line) and when the object begins to fall (right vertical line). Bottom - Corresponding internal pin positions from the same experiment as above. Inset in top left of each subfigure shows external tracking markers.}
    \label{fig:pins}
\end{figure}
%%%%%%%%%%%%%%%%%%%%%%%%%%%%%%%%%%%%%%%%%%%%%%%%%%%%%%%%%%%%
%%%%%%%%%%%%%%%%%%%%%%%%%%%%%%%%%
\begin{figure}[h]
    \centering
    \includegraphics[width=0.49\textwidth]{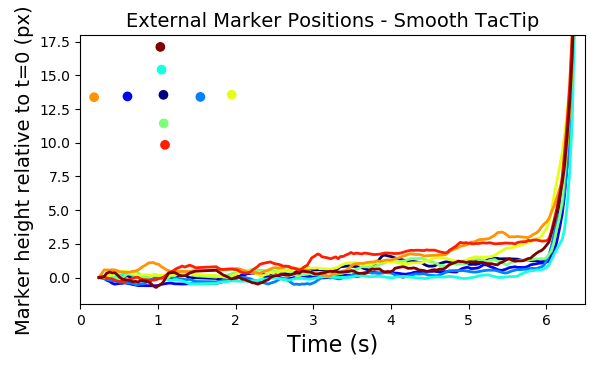}
    \caption{No incipient slip signal from external markers attached to smooth TacTip.}
    \label{fig:pins_old}
\end{figure}
%%%%%%%%%%%%%%%%%%%%%%%%%%%%%%%%
%%%%%%%%%%%%%%%%%%%%%%%%%%%%%%%%%%%%%%%%%%%%%%%%
% \begin{figure*}[t]
%     \centering
%     \begin{subfigure}[b]{0.32\textwidth}
%         \centering
%         \includegraphics[width=\textwidth]{figures/external_view_change.png}
%         \caption{}
%         \label{fig:pins_ext}
%     \end{subfigure}
%     \begin{subfigure}[b]{0.32\textwidth}
%         \centering
%         \includegraphics[width=\textwidth]{figures/internal_view_change.png}
%         \caption{}
%         \label{fig:pins_int}
%     \end{subfigure}
%     \begin{subfigure}[b]{0.32\textwidth}
%         \centering
%         \includegraphics[width=\textwidth]{figures/external_old_change.png}
%         \caption{}
%         \label{fig:pins_old}
%     \end{subfigure}
%     \caption{(a) External markers from newly designed sensor. Blue lines indicate where incipient slip begins (left vertical line) and when the object begins to fall (right vertical line), respectively. (b) Corresponding internal pin positions from the same experiment in (a). (c) External markers from the older TacTip design.  Inset in top left of each subfigure shows external tracking markers (for (a) and (c) as seen by external camera) and internal pins (for (b) as seen by internal camera)}
%     \label{fig:pins}
% \end{figure*}
%%%%%%%%%%%%%%%%%%%%%%%%%%%%%%%%%%%%%%%%%%%%%%%%%%%%%%%%%%%%
\subsection{Data Collection Rig} \label{sec:rig}
To collect data and perform a reliable comparison to the previous TacTip design we reuse the data collection rig from James et al. (2018) \cite{James2018SlipSensor}. The  apparatus  utilised  a  low  friction  rail  system  with a slider, to which objects can be attached (Fig. \ref{fig:rig}). This provides several advantages, including allowing: (i) objects to fall under gravity with minimal friction; (ii) experiments to be repeatable with high precision, and; (iii) experiments to be performed entirely without human input after initial setup.

Here, we augment the apparatus by adding an external camera to measure movement of the external surface of the sensor and the acrylic sheet. James et al. (2018) used various shapes made from ABS plastic; however, this is unsuitable here as we need to use a transparent object to make reference recordings of movements on the sensor's surface. We therefore use a sheet of 10 mm thick transparent acrylic. Additionally, we apply markers in a cross pattern to the surface of the sensor so that fixed points on the surface can be easily tracked by the external camera (Fig. \ref{fig:tips_side}). 

Both the external markers and the internal pins are identified and segmented using the Python OpenCV library function \textit{SimpleBlobDetection} which provides a list of blob $(x,y)$ coordinates for each frame. The order of the pins is not necessarily consistent between frames. However, by setting a threshold on the maximum distance a pin or marker can move between frames we can reorder the list of pins thus tracking them from frame to frame.

To annotate when gross slip occurs, we attach an ArUco optical marker to the acrylic so its height can be measured by the external camera. When the height of the acrylic sheet changes significantly, we assert that gross slip is occurring. To collect data the sensor is pressed against the acrylic, which is then lifted 50 mm. At this point, the cameras start to record and the arm retracts slowly (moving normal to the acrylic sheet) until the object has completely fallen. Different retraction speeds between $0.1$-$0.5$ mm.s$^{-1}$ at $0.1$ mm.s$^{-1}$ intervals are used.

\subsection{Change Point Detection} \label{sec:change_point}
To  label when incipient slip begins, we use the Pruned Exact Linear Time (PELT) algorithm \cite{Killick2012OptimalCost} applied to the internal pins. PELT is an example of a \textbf{change point} algorithm, the purpose of which is to calculate when the distribution of a data set changes. When the markers are static, the distribution of the detected positions will mostly consist of noise coming from the error inherent in the detection algorithm. When incipient slip occurs, we expect this distribution to shift and use change point detection to determine when this started. We choose PELT as it is known to be more accurate than other methods such as Binary Segmentation, and far more computationally efficient, executing in $\mathcal{O}(n)$ time. %Dorcas citation

\subsection{Deep Learning Architecture} \label{sec:dl}
To attempt the classification of incipient slip, we utilise a convolutional neural network (CNN). CNNs have shown strong performance in classification and regression using the TacTip, including estimating the orientation of edges \cite{Lepora2019FromSensor} and classifying objects \cite{Church2019TactileHand}. Using raw image data from the sensor has been shown to outperform using pin positions, as the network can learn from whichever features it deems are most informative rather than hand-crafted features. 

Here, we use a binary CNN to infer the presence of incipient slip on the \textbf{external} surface simply by observing the \textbf{internal} raw camera footage. Being a time-dependent phenomenon, incipient slip can only be detected by observing multiple frames of data. Therefore, we provide the network with multiple samples such that the differences in sensor data caused by incipient slip can be learned. We follow a similar image processing pipeline from previous work using the TacTip \cite{Lepora2019FromSensor, Church2019TactileHand, lepora2020optimal} which also reduces processing time: (i) Convert to grayscale; (ii) Apply Gaussian adaptive threshold using OpenCV function; (iii) Downsample image from (640$\times$480) to (55$\times$48) pixels to reduce input size to network; (iv) Sum the pixel values from ten consecutive frames to add a temporal dimension to the data; (v) Rescale the pixel values onto the domain [0,1] to reduce error gradients when training \cite{Bishop1995NeuralRecognition}. The network architecture consists of three convolutional layers with 32, 64 and 64 filters and max pooling between the first two layers, followed by a fully connected layer and an output layer.
%%%%%%%%%%%%%%%%%%%%%%%%%%%%%%%%%%%%%%%%%%%%%%%
\begin{figure}[t]
    \centering
    \begin{subfigure}[b]{0.48\textwidth}
        \includegraphics[trim={0 6cm 0 8cm}, clip=true, width=\textwidth]{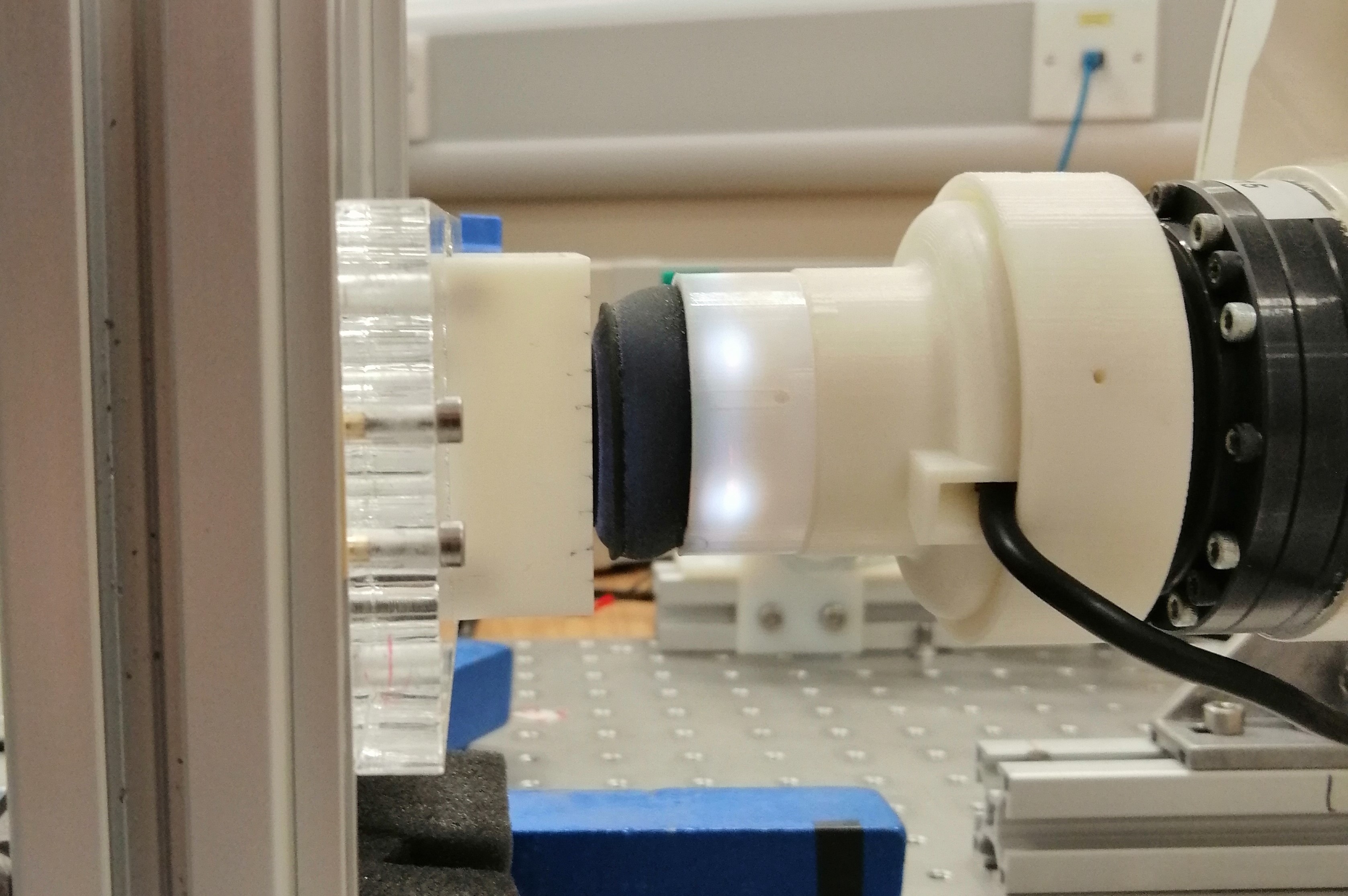}
        \label{fig:rig_flat_obj}
    \end{subfigure}
    \begin{subfigure}[b]{0.48\textwidth}
        \includegraphics[trim={0 1cm 0 0}, clip=true,width=\textwidth]{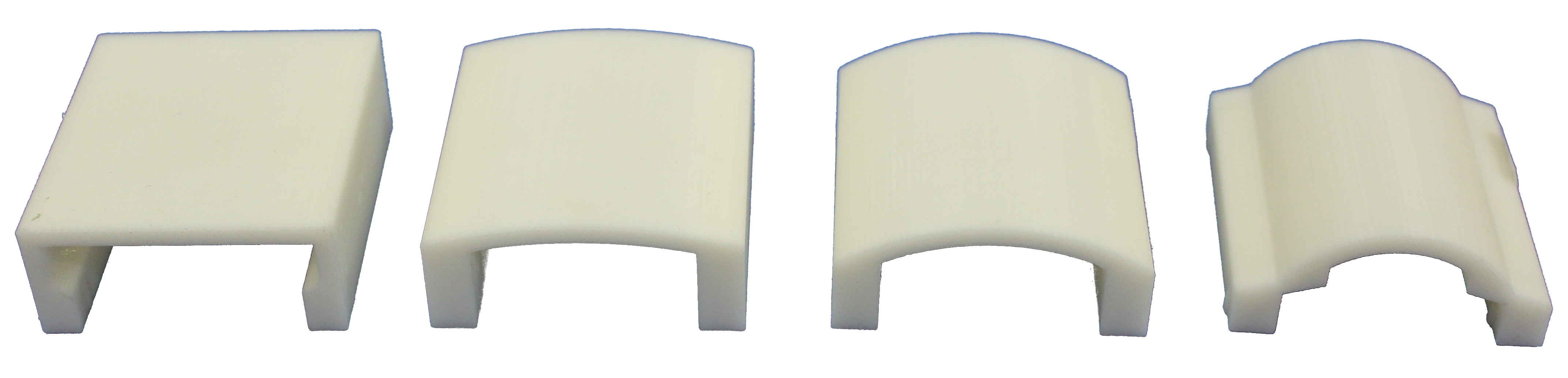}
        \label{fig:stimuli}
    \end{subfigure}
    \caption{Top: Flat stimulus held by the tip for real time testing. Bottom: Experiment stimuli with differing radii of curvature.}\label{fig:exp_online}
\end{figure}
%%%%%%%%%%%%%%%%%%%%%%%%%%%%%%%%%%%%%%%%%%%%%%%%%%%%%%%%%%%%%%%%%%%%%
\section{Results}
\subsection{Incipient Slip}
\subsubsection{External View Analysis}
Using the slip detection rig from Sec.\ref{sec:rig}, we performed 20 repeats at each retraction speed and recorded both the internal and external camera data. Following this, we detected the external markers and internal pins to determine whether incipient slip was present. Despite applying markers to the three innermost ridges, as well as the central pin, the third ring loses contact with the surface before gross slip occurs. Therefore, from here on, we ignored these four markers, leaving nine markers to be tracked, which were in contact for the duration of each trial.

The external camera video showed a clear presence of incipient slip (Fig. \ref{fig:tips_side}). The markers on the second ring displayed the most obvious movement however the marker atop this ring and the ArUco did not move. This means that parts of the contact surface were slipping when others were not and, crucially, before the object showed any sign of falling. This incipient slip pattern becomes clearer when looking at the data after the markers have been tracked. Again, it is clear that the farthest left and right markers begin to move after \textasciitilde2.5 s before any others (Fig.~\ref{fig:pins}). The markers inside the two markers on each extreme begin to move next. Throughout the experiment, the two topmost markers remain static until gross slip occurs and the object falls after \textasciitilde6.5 s.

The pattern seen here is broadly in line with predictions made by the Coulomb model of friction, which states that maximum static frictional force is proportional to the normal force on the surface. The domed shape means that the greater the radius of the ring, the lower the normal force, and therefore we would expect the markers further from the centre to slip first. When considering the line of markers placed horizontally across the middle of the sensor this is exactly what we see: the markers on the far left and far right slip first, followed by those markers closer to the centre. 

When considering the line of markers placed vertically up the middle of the sensor, the picture becomes more complicated. The top two markers show almost no motion, whereas the bottom two markers show noticeable movement. As the sensor is bearing the weight of the acrylic, this causes a large shear force to be applied to the sensor, which most likely causes the normal force distribution across the vertically-aligned markers to become asymmetric.

\subsubsection{Internal View Analysis}
There are two distinct types of sensor surface movement that cause the majority of the pin movement inside the sensor: translation and pivot. Translation occurs when an external ridge of the TacTip slips and the pin embedded within also translates in the same direction, but perhaps not to the same extent. Pivoting occurs when the ring surface remains static on the object but the parts of the skin spanning the ridges moves. As the pins protrude equal distances normal to the skin internally and externally, the effect of pivoting is quite pronounced. This effect somewhat confounds our ability to easily observe movement of the external surface of the ridge against the object (i.e. to detect slip).

To detect incipient slip on the inside we want to see examples of translational pin movement; however, it appears that the pivoting is the dominant effect resulting in pin movement. Thus, it is difficult to determine whether relative translation between pins relates to the occurrence of incipient slip on the external surface (Fig. \ref{fig:pins}), or whether this movement is due to pin pivoting. This does not mean that the effect is not detectable in the pin movement patterns, but rather that it is more challenging to extract. Fundamentally, this is a complex pattern recognition problem; thus, we are motivated to utilise a deep learning approach directly on the tactile images.  

\subsubsection{TacTip Marker Analysis}
To confirm that incipient slip is not visible in the previous smooth TacTip design used in previous work~\cite{James2018SlipSensor}, we attached markers in the same pattern on the surface of that smooth TacTip and performed the same experiment. As expected, the movement of markers does not display the characteristic incipient slip pattern that is present in the new design (Fig. \ref{fig:pins_old}). Instead, all markers remain static until gross slip occurs, when they all move in unison. This bulk motion of the surface contributed to making that smooth design of TacTip so effective at detecting gross slip \cite{James2018SlipSensor}.

\subsection{Online Testing}
\subsubsection{Data Labelling}
To detect the time that incipient slip commences we use the change point detection algorithm PELT (Sec. \ref{sec:change_point}). Since PELT requires the entire time series of data to determine the change points, it cannot be used in real time to detect changes. We use it to detect the onset of incipient slip by providing the data from the two external markers which exhibit the strongest incipient slip signal (Fig. \ref{fig:tips_side}). The detected change point is used as the label to define when incipient slip begins. We use the ArUco marker recording to determine when gross slip begins. As the classifier is binary labels are either \textit{incipient slip} or \textit{not}.
%%%%%%%%%%%%%%%%%%%%%%%%%%%%%%%%
\begin{table}[b]
\begin{center}
\begin{tabular} {p{1.5cm } p{1.4cm} p{3.2cm}}
\hline \hline
RoC (mm) & Success (\%) &  Incipient Slip Margin (mm)  \\ \hline \hline
\(\infty\) (Flat) & 100 & 0.9 \\ \hline
80 & 100 & 0.9 \\ \hline
40 & 100 & 1.0 \\ \hline
20 & 100 & 0.7 \\
\hline \hline
\end{tabular}
\end{center} 
\caption{The success of incipient slip detection with objects with varying radii of curvature (RoC). Twenty tests were performed per object.}
\label{tab:online}
%\vspace{-1em}
\end{table}
%%%%%%%%%%%%%%%%%%%%%%%%%%%%%
\subsubsection{Convolutional Neural Network} We use 50\% of the labelled data for training the CNN, withholding 25\% for each of the validation and testing sets to judge performance. We ensure that an equal proportion of experiments from each retraction speed is contained within each data set to avoid training bias. Additionally, we augment the training data set by performing random shifting of the image by 20\% in $x$ and $y$ and randomly zooming in on the image by up to 20\%. This was performed to help the network generalise to changes in shear and normal forces caused by varying objects. We trained for ten epochs, after which the validation error began to increase, thus avoiding overfitting the data.

After training we quantify how successful the model is at detecting incipient slip on the unseen test data set. The model predicted whether incipient slip was present or not with over 98\% accuracy. This suggests that the model has not overfitted to the training data and is highly successful at identifying the presence of incipient slip using the internal camera. However, the real test of the model's ability to detect incipient slip is to use it in real time on different objects.  

\subsubsection{Experiment Description}
To test incipient slip detection in real time, we reuse four tactile stimuli from James et al. (2018) (Fig. \ref{fig:exp_online}). Each has a different radius of curvature that will provide a different normal force distribution across the sensor surface. They are also opaque so the contact surface cannot be observed. When testing online we want to challenge the incipient slip CNN by varying the conditions under which incipient slip occurs. As these stimuli change the object weight, texture (which in turn can alter the frictional properties by affecting real contact area between the sensor and object), and normal force distribution, they provide an indication of how well the model has generalised.
%%%%%%%%%%%%%%%%%%%%%%%%%%%%%%%%%
\begin{figure}[t]
    \centering
    \includegraphics[trim={0 0.2cm 0 0}, clip=true, width=0.85\linewidth]{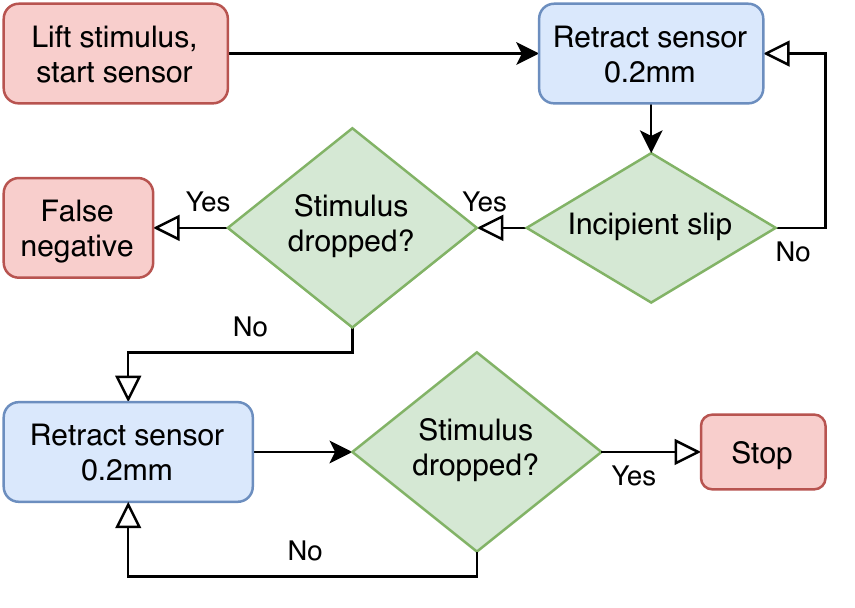}
    \caption{Experiment flowchart.}
    \label{fig:flow}
\end{figure}
%%%%%%%%%%%%%%%%%%%%%%%%%%%%%%%%

In the specific experiment shown in Fig.~\ref{fig:pins}, the sensor was retracted at 0.2 mm.s$^{-1}$, and incipient slip commences approximately four seconds prior to gross slip. This means that we have a buffer zone of nearly $0.8$ mm of movement where the sensor indicates that the object is at risk of experiencing gross slip. To test how far prior to gross slip our model can detect incipient slip, we follow the procedure in Fig. \ref{fig:flow}. After incipient slip is detected, we can count how many additional 0.2 mm retractions occur before the object falls, and determine how far in advance of gross slip we are detecting incipient slip. We repeated the experiment twenty times per stimulus with a randomly chosen retraction speed between 0.1-0.5 mm.s$^{-1}$. Results are presented in Table \ref{tab:online}.

In all trials for each stimulus, incipient slip was detected before the object fell. The most curved stimulus (20mm radius of curvature) had the smallest incipient slip margin (0.7 mm), meaning that it was closest to slipping when incipient slip was detected. The remaining three stimuli had broadly similar margins between 0.9-1.0 mm.  Therefore in all cases, incipient slip was detected considerably before the object fell, with a known zone of safety, which could be used to proactively prevent slippage of a held object.

\section{Discussion}
In this paper, we presented a modified TacTip biomimetic optical tactile sensor that can facilitate and detect incipient slip, as confirmed by recording the movement of markers on the external surface of the sensor. This is significant as we conclusively show the presence of incipient slip whereas many prior studies assert its existence from data \cite{su2015force,Fujimoto,Veiga2015StabilizingSlip}. Several changes to the TacTip design were made based on the physical conditions which facilitate incipient slip. The most significant change involved the addition of ridges - structurally analogous to human the fingerprint - in a concentric ring pattern which allowed for localised shear deformation to occur on the sensor surface for a significant duration prior to the onset of gross slip.

When testing on unseen objects, the detection of incipient slip indicated that gross slippage of the object would occur were the sensor to be retracted normally from the test object between 0.6 and 1.0 mm. These results are in accordance with another experiment considered here where we directly observed the movement of external markers indicating incipient slip, with 0.8mm of retraction distance before gross slip occurs (Fig.\ref{fig:pins}). Using incipient slip as an indicator that one's grasp on an object is insecure is a far more effective strategy than waiting for gross slip to occur and then attempting regrasp the object and arrest the gross slip, as previously performed with the TacTip \cite{James2018SlipSensor}. In practice, this could give a sizeable time window of at least 1.2s, when moving at 0.5 mm.s$^{-1}$, during which to take corrective action to prevent gross slip. This is larger than the slip prediction windows of 30 ms achieved by Su et al. (2015) \cite{su2015force} and 20 ms reported by Veiga et al. \cite{Veiga2015StabilizingSlip}. However, here we are using a very limited set of objects, so further work is needed for a complete comparison. Additionally, it may not be necessary to react immediately upon incipient slip onset so providing a metric of proximity to gross slip onset would be useful.

A limitation of the new design is that the incipient slip signal pattern contained in the movement of pins on the inside of the tactile sensor is not amenable to straightforward human interpretation. This meant that we resorted to utilising deep learning methods to detect the internal incipient slip signal, from which we obtained over 98\% classification accuracy. Further work with a more diverse training set and testing on a wider array of objects will be needed to determine how well the CNN can detect incipient slip. In addition, future work will involve further modifications of the design such that the internal pins movements more directly correspond to movements on the external surface, thereby clearly exhibiting the relative shear displacements associated with incipient slip. This could include adding ridges to the inside of the skin and making the pins shorter to reduce pivoting. Other future work could be to vary the size and density of rings so that they more closely resemble the human fingerprint although we are limited by 3D printing resolution and ability to resolve markers in the rings with a camera.

To summarise, we have developed a tactile sensor which can reliably detect and facilitate incipient slip. This is of direct practical utility to robotic hand research on dexterous manipulation of a grasped objects, as it would greatly reduces the chance of manipulated objects being dropped. 

\section*{Acknowledgements}

The authors thank members of the BRL Tactile Robotics Group for their help and insights throughout this work.

\bibliographystyle{IEEEtran}
\bibliography{references_cut.bib}

\end{document}